\documentclass[a4paper,twoside]{article}

\usepackage{epsfig}
\usepackage{subcaption}
\usepackage{calc}
\usepackage{amssymb}
\usepackage{amstext}
\usepackage{amsmath}
\usepackage{amsthm}
\usepackage{multicol}
\usepackage{pslatex}
\usepackage{apalike}
\usepackage{wasysym}
\usepackage{multirow, subcaption}
\usepackage{xr}
\usepackage{SCITEPRESS}     

\begin{document}
\title{Improving the Sample-Complexity of Deep Classification Networks with Invariant Integration}

\author{\authorname{Matthias Rath\sup{1,2}, Alexandru Paul Condurache\sup{1,2}}
\affiliation{\sup{1}Automated Driving Research, Robert Bosch GmbH, Stuttgart, Germany}
\affiliation{\sup{2}Institute for Signal Processing, University of L\"ubeck, L\"ubeck, Germany}
\email{\{Matthias.Rath, AlexandruPaul.Condurache\}@de.bosch.com}
}


\abstract{Leveraging prior knowledge on intraclass variance due to transformations is a powerful method to improve the sample complexity of deep neural networks. This makes them applicable to practically important use-cases where training data is scarce. Rather than being learned, this knowledge can be embedded by enforcing invariance to those transformations. Invariance can be imposed using group-equivariant convolutions followed by a pooling operation.\\
For rotation-invariance, previous work investigated replacing the spatial pooling operation with invariant integration which explicitly constructs invariant representations. Invariant integration uses monomials which are selected using an iterative approach requiring expensive pre-training. We propose a novel monomial selection algorithm based on pruning methods to allow an application to more complex problems. Additionally, we replace monomials with different functions such as weighted sums, multi-layer perceptrons and self-attention, thereby streamlining the training of invariant-integration-based architectures.\\
We demonstrate the improved sample complexity on the Rotated-MNIST, SVHN and CIFAR-10 datasets where rotation-invariant-integration-based Wide-ResNet architectures using monomials and weighted sums outperform the respective baselines in the limited sample regime. We achieve state-of-the-art results using full data on Rotated-MNIST and SVHN where rotation is a main source of intraclass variation. On STL-10 we outperform a standard and a rotation-equivariant convolutional neural network using pooling.
}

\onecolumn \maketitle \normalsize \setcounter{footnote}{0} \vfill

\section{INTRODUCTION}
Deep neural networks (DNNs) excel in problem settings where large amounts of data are available such as computer vision, speech recognition or machine translation \cite{DeepLearning}.
However, in many if not most real-world problem settings training data is scarce because it is expensive to collect, store and in case of supervised training label. Consequently, an important aspect of DNN research is to improve the sample complexity of the training process, i.e., achieving best results when the available training data is limited.

One solution to reduce the sample complexity is to incorporate meaningful \textit{prior knowledge} to bias the learning mechanism and reduce the complexity of the possible parameter search space. One well-known example on how to embed \textit{prior knowledge} are convolutional neural networks (CNNs) which achieve state-of-the-art performance in a variety of tasks related to computer vision. CNNs successfully employ translational weight-tying such that a translation of the input leads to a translation of the resulting feature space.
This property is called \textit{translation equivariance}.

These concepts can be expanded such that they cover other transformations of the input which lead to a predictable change of the output -- or to no change at all. The former is called \textit{equivariance} while the latter is a related concept referred to as \textit{invariance}.

In general, DNNs for image-based object detection and classification hierarchically learn a set of features that ideally contain all relevant information to distinguish different objects while dismissing the irrelevant information contained in the input.
Generally, transformations causing intraclass variance can act globally on the entire input image, e.g., global rotations or illumination changes -- or locally on the objects, e.g., perspective changes, local rotations or occlusions.
Prior knowledge about those transformations can usually be obtained before training a DNN and thus be incorporated to the training process or architecture. Enforcing meaningful \textit{invariances} on the learned features simplifies distinguishing relevant from irrelevant input information.
One method to enforce invariance is to approximately learn it via \textit{data augmentation}, i.e., artificially transforming the input during training. However, these learned invariances are not exact and do not cover all relevant variability. 

\cite{GroupEquivariantCNNs} first applied \textit{group-equivariant convolutions} (G-Convs) to DNNs. G-Convs mathematically guarantee equivariance to transformations which can be modeled 
as a group. A DNN consisting of multiple layers is equivariant with respect to a transformation group, if each of its layers is group-equivariant or commutes with the group \cite{GroupEquivariantCNNs}. Consequently, an equivariant DNN consists of multiple group-convolutional layers as well as pooling and normalization operations that commute with the desired transformations. For a classifier, the G-Convs are usually followed by a global pooling operation over both the group dimension and the spatial domain in order to enforce invariance. These invariant features are then processed by fully connected layers to obtain the final class scores. 

Invariant Integration (II) is a method to explicitly create a complete, invariant feature space with respect to a transformation group introduced by \cite{SM_Existence,SM_Algos}. Recent work showed that explicitly enforcing rotation-invariance by means of II instead of using a global pooling operation among the spatial dimensions decreases the sample complexity of rotation-equivariant CNNs used for classification tasks despite adding parameters, hence improving generalization \cite{Rath}. However, II thus far relies on calculating monomials which are hard to optimize with usual DNN training methods. Additionally, monomial parameters have to be chosen using an iterative method based on the least square error of a linear classifier before the DNN can be trained. This method relies on an expensive pre-training step that reduces the applicability of II to real-world problems. 

Consequently, in this paper we investigate how to adapt the rotation-II framework in combination with equivariant backbone layers in order to reduce the sample complexity of DNNs on various real-world datasets while simplifying the training process. Thereby, we explicitly investigate the transition between in- and equivariant features for the case of rotations and replace the spatial pooling operation by II. We start by introducing a novel monomial selection algorithm based on pruning methods. Additionally, we investigate replacing monomials altogether, using simple, well-known DNN layers such as a weighted sum (WS), a multi-layer perceptron (MLP) or self-attention (SA) instead. This contributes significantly to streamlining the entire framework. We specifically apply these approaches to 2D rotation-invariance. We achieve state-of-the-art results irrespective of limited- or full-data regime, when rotations are responsible for most of the relevant variability, such as on Rotated-MNIST and SVHN. Furthermore, we demonstrate very good performance in limited-data regimes on CIFAR-10 and STL-10, when besides rotations also other modes of intraclass variation are present.

\begin{figure*}[t]
	\centering
	\includegraphics[width=0.7\linewidth]{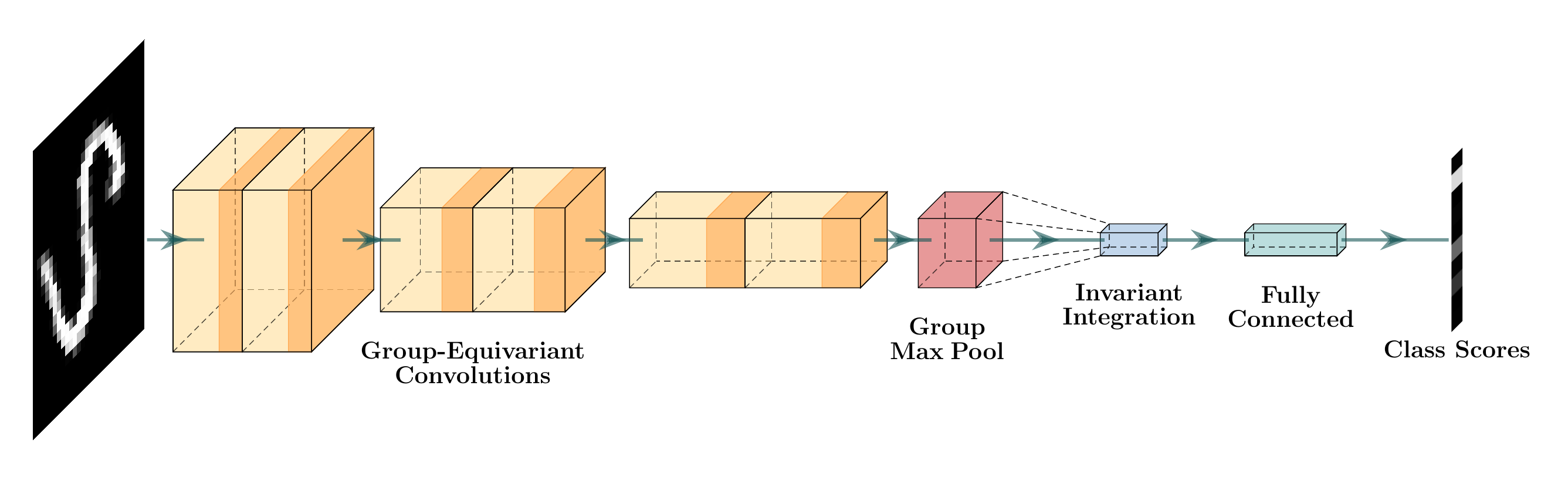}
	\caption{General invariant DNN architecture using II. The architecture includes group convolutions (orange) and group pooling (red) creating an equivariant representation, an II layer enforcing invariance (blue), and fully-connected layers (green).}\label{fig:Architecture}
\end{figure*}

Our \textbf{core contributions} are:
\begin{itemize}
	\itemsep 0em
	\item We introduce a novel algorithm for the II monomial selection based on pruning. 
	\item We investigate various functions to replace the monomials within the II framework including a weighted sum, a MLP and self-attention. We thereby streamline the training process of II-enhanced DNNs as the \textit{monomial selection is no longer needed}.
	\item We demonstrate the performance of rotation-II on the real world datasets SVHN, CIFAR-10 and STL-10.
	\item We apply II to Wide-ResNet (WRN) architectures, demonstrating its general applicability. 
	\item We establish a connection between II and regular G-Convs.
	\item We show that using II in combination with equivariant G-Convs reduces the sample complexity of DNNs.
\end{itemize} 

\section{RELATED WORK}
DNNs can learn invariant representations using \textbf{group-equivariant convolutions} or \textbf{equivariant attention} in combination with pooling operations. Other methods explicitly \textbf{learn invariance}, or enforce it using \textbf{invariant integration}.

\textbf{Group-equivariant convolutional} neural networks (G-CNNs) are a general framework to introduce equivariance, first proposed and applied to $90^\circ$ rotations and flips on 2D images by \cite{GroupEquivariantCNNs}. G-CNNs were extended to more fine-grained or continuous 2D rotations \cite{HarmonicNetworks,Bekkers,PCAM,SFCNN,Winkels,Diaconu1,ECCO}, processed as vector fields \cite{MarcosRot} or further generalized to the $\text{E}(2)$-group which includes rotations, translations and flips \cite{E2STCNNs}. Additionally, 2D scale-equivariant group convolutions have been introduced \cite{XuScale,LocalScaleInvariance,MarcosScale,GhoshScale,ZhuScale,Worrall19,Sosnovik}. Further advances include expansions towards three-dimensional spaces (e.g., \cite{CubeNet,ClebschGordanNets,Esteves}) or general manifolds and groups (e.g., \cite{Gauge,GeneralEquivariantCNNs,BekkersLie,Finzi2021}) which are beyond the scope of this paper.

Recently, \textbf{equivariance} was also introduced to \textbf{attention} layers. 
\cite{Diaconu2,RomeroCoAttentive,AttentiveGCNN} combined equivariant attention with convolution layers to enhance their expressiveness. \cite{SE3Transformer,IterativeSE3Transformer,G-SelfAttention,LieTransformer} introduced different equivariant transformer architectures.
In order to obtain invariant representations, equivariant layers are usually combined with pooling operations. 

Other methods to \textbf{learn invariant representations} include data augmentation, pooling over all transformed inputs \cite{Laptev}, learning to transform the input or feature spaces to their canonical representation \cite{STN,PolarTN,ETN} or regularization methods \cite{RegularizedInvariance}. However, these methods approximate invariance rather than enforcing it mathematically guaranteed. 

\textbf{Invariant integration} is a principled method to enforce invariance. It was introduced as a general algorithm in \cite{SM_Existence,SM_Algos} and applied in combination with classical machine learning classifiers for various tasks such as rotation-invariant image classification \cite{SM_Gray}, speech recognition \cite{Muller1,Muller2,Muller3}, 3D-volume and -surface classification \cite{II3d} or event detection invariant to anthropometric changes \cite{Condurache}. 

In \cite{Rath}, rotation-II was applied in combination with steerable G-Convs in DNNs for image classification. The equivariant feature space learned by the G-Convs is followed by max-pooling among the group elements. Rotations of the input induce rotations in the resulting feature space, i.e., it is equivariant to rotations. While standard G-CNNs employ spatial max-pooling afterwards to achieve an invariant representation, \cite{Rath} and our approach replace it with II, which increases the expressibility compared to spatial max-pooling while still guaranteeing invariant features. These are finally processed with dense layers to calculate the classification scores (see Figure \ref{fig:Architecture}). 

All previous methods including \cite{Rath} used II in combination with monomials which were either hand-designed or selected using expensive iterative approaches which required pre-training the entire network without II. In contrast, we propose a novel pre-selection algorithm based on pruning methods or to replace the monomials altogether. Both approaches can be applied to DNNs more natively.

\section{PRELIMINARIES}
In this section we concisely present the mathematical principles needed to define in- and equivariance in DNNs which rely on Group Theory. Furthermore, we introduce group-equivariant convolutions which are used to obtain equivariant features and form the backbone of our DNNs.
\subsection{In- \& Equivariance}
A group $G$ is a mathematical abstraction consisting of a set $\mathcal{X}$ and a \textit{group operation} $\cdot : G \times G \rightarrow G$ that combines two elements to form a third. A group fulfills the four axioms closure, associativity, invertibility and identity. Group Theory is important for DNN research, because invertible transformations acting on feature spaces can be modeled as a group, where the left group action
$G \times \mathbb{R}^n \rightarrow \mathbb{R}^n, (g,x) \mapsto L_g x$ with $g \in G$ acts on the vector space $\mathbb{R}^{n}$.

The concept of \textit{in-} and \textit{equivariance} can be mathematically defined on groups. A function $f : \mathbb{R}^n \rightarrow \mathbb{R}^m$ is defined as \textit{equivariant}, if its output $f(x)$ transforms predictably under group transformations
\begin{equation}
\forall \; g \; \exists \; g^\prime \; \text{s.t.} \; f(L_gx) = L_{g^\prime} f(x) \text{,}
\end{equation}
for all $x \in \mathbb{R}^n$,
and $g \in G, g^\prime \in G^\prime$
while $G \text{ and } G^\prime$ may be the same or different groups. 
If the output does not change under transformations of the input, i.e., $\forall \; g \; \forall \; x, \; f(L_gx)=f(x)$, $f$ is \textit{invariant} \cite{GroupEquivariantCNNs}.

\subsection{Group-Equivariant Convolutions}
\cite{GroupEquivariantCNNs} first used the generalization of the convolution towards general transformation groups $G$ in the context of CNNs. The discrete group-equivariant convolution of a signal $x$ and a filter $\psi: G \rightarrow \mathbb{R}^{n}$ is defined as
\begin{equation}
[x \star_G \psi](g) = \sum_{h\in G}x(h)\psi(g^{-1}h) \text{.}
\end{equation}
Here, $x : G \rightarrow \mathbb{R}^{n}$ is used as a function. Both definitions are interchangeable.
The standard convolution is a special case where $G=\mathbb{Z}^2$.
The output of the group-convolution is no longer defined on the regular grid, but on group elements $g$ and is equivariant w.r.t $G$. The action $L_{g^\prime}$ in the output space depends on the group representation that is used. Two common representations used for G-CNNs are the \textit{irreducible representation} and the \textit{regular representation} which consists of one additional \textit{group channel} per group element storing the responses to all transformed versions of the filters. Often, the regular representation is combined with \textit{transformation-steerable filters} which can be transformed arbitrarily via a linear combination of basis filters, hence avoiding interpolation artifacts \cite{Adelson,SFCNN,3DSTCNNs,E2STCNNs,GhoshScale,Sosnovik}.
In order to obtain invariant features from the equivariant ones learned by G-CNNs, a pooling operation is usually employed.

\section{INVARIANT INTEGRATION}
Invariant Integration introduced by \cite{SM_Existence} is a method to create a complete feature space w.r.t. a group transformation based on the Group Average $A[f](x)$ 
\begin{equation}\label{eq:GroupAverage}
A[f](x) = \int_{g\in G} f(L_gx) d\mu(g) \text{,}
\end{equation}
where $\int d\mu(g)=1$ defines the Haar Measure and $f$ is an arbitrary complex-valued function. A complete feature space implies that all patterns that are equivalent w.r.t $G$ are mapped to the same point while all non-equivalent patterns are mapped to distinct points.

\subsection{Monomials}
For the choice of $f$, \cite{SM_Existence} proposes to use the set of all possible monomials which form a finite basis of the signal space according to \cite{Noether1916}. Monomials are a multiplicative combination of different scalar input values $x_i$ with exponents $b_i \in \mathbb{R}$
\begin{equation}
m(x) = \prod_{i=1}^{M}x_i^{b_i} \quad \text{with} \quad \sum_{i=1}^{M}b_i \leq|G|\text{.}
\end{equation}
Combined with the finite group average, we obtain
\begin{equation}
A[m](x) = \frac{1}{|G|}\sum_{g\in G} m(L_gx) = \frac{1}{|G|}\sum_{g\in G} \prod_{i=1}^{M}(L_gx)_i^{b_i} \text{,}
\end{equation}
\cite{SM_Algos}.
When applying II with monomials to two-dimensional input data on a regular grid such as images, it is straightforward to use pixels and their neighbors for the monomial factors $x_i$. Consequently, monomials can be defined via the distance of the neighbor to the center pixel $d_i$
with $d_1=0$. For discrete 2D rotations and translations, this results in the following formula \cite{SM_Gray}
\begin{equation}
\small
A[m](x) = \frac{1}{UV\Phi}\sum_{u,v,\phi} \prod_{i=1}^{M}x[u+\cos(\phi)d_i,v+\sin(\phi)d_i]^{b_i} \textit{,}
\end{equation}
which can be used within a DNN with learnable exponents $b_i$ \cite{Rath}.

\subsection{Monomial Selection}\label{sec:MonomialSel}
While the II layer reduces the sample complexity when learning invariant representations, it introduces additional parameters which need to be carefully designed. One of them is the selection of a meaningful set of parameters $d_i$ and $b_i$ that define the monomials needed to obtain the invariant representation. This step is necessary since the number of possible monomials satisfying $\sum_i b_i \leq|G|$ is too extensive. 

In \cite{Rath}, an iterative approach is used based on the least square error solution of a linear classifier. While the linear classifier is easy to compute, the iterative selection is time-consuming and computationally expensive. Additionally, the base network without the II layer needs to be pre-trained which requires additional computations and prevents training the full network from scratch.

Consequently, we investigate alternative approaches for the monomial selection. Two selection approaches are introduced and explained in the following. Both enable training the network end-to-end from scratch and are computationally inexpensive compared to the iterative approach.

\subsubsection{Random Selection}
First, we randomly select the $n_m$ monomials by sampling both the exponents and the distances from uniform distributions. This approach is fast only requiring a single random sampling operation and serves as a baseline to evaluate other selection methods.

\subsubsection{Pruning Selection}
Alternatively, monomial selection can be formulated as selecting a subset containing $n_m \ll M$ possible monomial parameters. Consequently, it is closely related to the field of pruning in DNNs whose goal is to reduce the amount of connections or neurons within DNN architectures in order to reduce the computational complexity while maintaining the best possible performance.
We compare two pruning algorithms: a magnitude- and a connectivity-based approach.

\medskip
\noindent\textbf{Magnitude-Based Approach} ~
\cite{MagnitudePruning} determine the importance of connections in DNNs by pre-training the network for $\tau$ epochs and sorting the weights of all layers by their magnitude $|w_{ij}|$. This approach is applied iteratively keeping the $\gamma$ highest-ranked connections at each step until the final pruning-ratio $\gamma^I$ is reached. 

Since we aim to prune monomials instead of single connections, we sum the absolute value of weights connected to a single monomial, i.e., all weights of the first fully connected layer following the II layer:
\begin{equation}
s_j = \frac{1}{C_iC_o}\sum_{k=1}^{C_i}\sum_{l=1}^{C_o}|w_{klj}| \text{,}
\end{equation}
where $C_i$ is the number of input channels before II is applied, $C_o$ is the number of neurons in the fully connected layer and $j$ selects the connections belonging to the $j^{\text{th}}$ monomial. 
Following \cite{MagnitudePruning}, we apply the pruning iteratively. In each step, we keep the $n_i$ monomials with the highest calculated score $s_j$. We do not re-initialize our network randomly in between iterative steps but re-load the pre-trained weights from the previous step. 

\medskip
\noindent\textbf{Connectivity-Based Approach} ~
We examine a second pruning approach based on the initial connectivity of weights inspired by \cite{SNIP}. All monomial output connections are multiplied with an indicator mask $c \in \{0, 1\}^{M}$ using the Hadamard product $c \odot w_{l}$. Here, $w_l$ is the weight vector of the fully connected layer following the II step. Setting an individual value $c_j$ to zero results in deleting all connections $w_{l,j}$ connected to monomial $j$. 
Consequently, the effect of deactivating a monomial can be estimated w.r.t the training loss $L$ by calculating the connection sensitivity
\begin{equation}
s_j = \frac{\partial L(c,w_l;\mathcal{D})}{\partial c_j}
\end{equation}
for each monomial using backpropagation. The $n_m$ monomials with the highest connection sensitivity are kept. The derivative is calculated using the training dataset $\mathcal{D}$.

The connectivity-based approach can either be used directly after initializing the DNN, or after some pre-training steps. Additionally, it can either be used iteratively or in a single step.

\subsubsection{Initial Selection}
We investigate two different approaches for the initial selection of $M \gg n_m$ monomials. In addition to a purely random selection, we design a catalog-based initial selection in which all possible distance combinations are guaranteed to be involved in the initial set. In both cases, we sample the exponents randomly from a uniform distribution.

\subsection{Replacing the Monomials}
In addition to the novel monomial selection algorithm, we investigate alternatives for the monomials used to calculate the group average (Equation \ref{eq:GroupAverage}). We apply the proposed functions to the group of discrete 2D rotations and compare monomials to well-utilized DNN functions such as a weighted sum, a MLP and a self-attention-based approach.

\subsubsection{Weighted Sum}
One possibility for $f$ is to use a weighted sum where the weights are a learnable kernel $\psi$ applied at each group element $g$ transforming the input $x$. We obtain
\begin{align}\label{eq:II_Weighted}
A[\text{WS}](x) 
= \frac{1}{|G|}\sum_{g\in G}\sum_{y \in \mathbb{Z}^2} x(y)\psi(g^{-1}y) \text{.}
\end{align}
For 2D-rotations, this results in translating and rotating the kernel using all group elements $g \in \mathrm{SO}(2)$.
We implement two different versions of II with WS. First, we apply a global convolutional filter, i.e., the kernel size is equivalent to the size of the input feature map (\textit{Global-WS}). Secondly, we use local filters with kernel size $k$ which we apply at all spatial locations $(u,v)$ and all orientations $\phi$ (\textit{Local-WS}).

\medskip
\noindent\textbf{Relation to Group Convolutions} ~
In the following we show the close connection between II using a WS and the group convolution introduced by \cite{GroupEquivariantCNNs}. Recall the formulation of the discrete group convolution of an image $x: \mathbb{Z}^2 \rightarrow \mathbb{R}$ and a filter $\psi: \mathbb{Z}^{k \times k} \rightarrow \mathbb{R}$ 
\begin{equation}
[x\star \psi](g) = \sum_{y\in \mathbb{Z}^2} x(y) \psi(g^{-1}y) \text{.} 
\end{equation}

The group convolution followed by global average pooling $A_G\{\cdot\}$ among all group elements is 
\begin{equation}
A_G\{[x\star \psi](\cdot)\} =
\frac{1}{|G|}\sum_{g \in G}\sum_{y\in \mathbb{Z}^2} x(y) \psi(g^{-1}y) \text{,}
\end{equation}
which is exactly the same formulation as Equation \ref{eq:II_Weighted}.
Thus, using a regular lifting convolution and applying global average pooling can be formulated as a special case of II.

\subsubsection{Multi-Layer Perceptron}
Another possibility for $f$ is a multi-layer perceptron (MLP) which consists of multiple linear layers and non-linearities $\sigma$. In combination with the rotation-group average, we obtain
\begin{equation}
\small
A[\text{MLP}](x) = \frac{1}{UV\Phi}\sum_{u,v,\phi}\sigma(\mathbf{W}_l \cdot \cdot \cdot \sigma(\mathbf{W_1}L_{g_\phi}^{-1}x_{\mathcal{N}}))\text{,}
\end{equation}
where $\mathcal{N}$ defines the neighborhood of a pixel located at $(u,v)$. For $\sigma$, we choose to use ReLU non-linearities.

\subsubsection{Self-Attention}
Finally, we insert a self-attention module into the II framework. Visual self-attention $\text{SA}(\mathbf{x})$ is calculated by defining the pixels of the input image or feature space $x$ as $N=H\cdot W$ individual tokens $\mathbf{x}\in\mathbb{R}^{HW\times C_i}$ with $C_i$ values and learning attention scores $\mathbf{A} \in \mathbb{R}^{N\times N}$. It includes three learnable matrices: the value matrix $\mathbf{W}_V \in \mathbb{R}^{C_i \times C_h}$, the key matrix $\mathbf{W}_K \in \mathbb{R}^{C_i \times C_h}$ and the query matrix $\mathbf{W}_Q \in \mathbb{R}^{C_i \times C_h}$. It is defined as
\begin{equation}
\small
\text{SA}(\mathbf{x}) = \text{softmax}(\mathbf{A})\mathbf{x}\mathbf{W}_{V}
\; \text{with}
\; \mathbf{A} = \mathbf{x}\mathbf{W}_Q(\mathbf{x}\mathbf{W}_K)^T \text{.}
\end{equation}
To incorporate positional information between the individual pixels, we use relative encodings $\mathbf{P}$ between query pixel $x_i$ and key pixel $x_j$ \cite{PeterShaw}
\begin{equation}
\mathbf{A_{i,j}} = \mathbf{x_i}\mathbf{W}_Q((\mathbf{x_j}+\mathbf{P}_{x_j-x_i})\mathbf{W}_K)^T \text{.}
\end{equation}
We embed this formulation into the II framework by transforming the input using bi-linear interpolation and apply the group average over all results. 

\begin{equation}
A[\text{SA}](x) = \frac{1}{|G|}\sum_{g \in G}\text{softmax}(L_g\mathbf{A})L_g\mathbf{x}\mathbf{W}_{V} \text{,}
\end{equation}
where $L_g\mathbf{A}$ denotes calculating the attention scores using the transformed input. We also investigate multi-head self-attention (MH-SA) where $H$ self-attention layers are calculated, concatenated and processed by a linear layer with weights $\mathbf{W}_{o} \in \mathbb{R}^{HC_h\times C_o}$. This formulation is related to \cite{G-SelfAttention}, where opposed to our approach equivariance is enforced using adapted positional encodings.

\section{EXPERIMENTS \& DISCUSSION}
We evaluate the different setups on Rotated-MNIST, SVHN, CIFAR-10 and STL-10. For each dataset, we choose a baseline architecture, assume that the feature extraction network is highly optimized and focus on the role of the II layer. We keep the number of parameters for the equivariant networks constant by adapting the number of channels per layer (see Appendix). We conduct experiments using the full training data, but more importantly limited subsets to investigate the sample complexity of the different variants. When training on limited datasets, we keep the number of total training iterations constant and adapt all hyper-parameters depending on epochs, such as learning rate decay, accordingly. All data subsets are sampled randomly with constant class ratios and are equal among all architectures. 
We optimized the hyper-parameters using Bayesian Optimization with Hyperband \cite{BOHB} and a train-validation split of 80/20.
Implementation details and hyper-parameters can be found in the Appendix.

\subsection{Evaluating Monomial Selection}
\begin{table}[t]
	\centering
	\small
	\caption{Mean Test Error (MTE) of different monomial selection types on Rotated-MNIST using II-SF-CNN \cite{Rath}. $\checkmark$ indicates full pre-training, $\LEFTcircle$ iterative pre-training for a small number of epochs and x pruning at initialization.}\label{tab:MonomialSelection}
	\begin{tabular}{lccc}
		\hline
		Selection & Pre-Train & Init. & MTE [\%] \\
		\hline
		SF-CNN & - & - & 0.714 $\pm$ 0.022 \\
		\hline
		- & x & Random & 0.751 $\pm$ 0.032 \\
		LSE & \checkmark & Random & 0.687 $\pm$ 0.012 \\
		\hline
		Connectivity & x & Random & 0.758 $\pm$ 0.0025 \\
		Connectivity & $\LEFTcircle$ & Catalog & 0.708 $\pm$ 0.010 \\
		Connectivity & $\LEFTcircle$ & Random & 0.705 $\pm$ 0.027 \\
		Magnitude & $\LEFTcircle$ & Catalog & 0.704 $\pm$ 0.022 \\
		Magnitude & $\LEFTcircle$ & Random & 0.677 $\pm$ 0.031 \\
	\end{tabular}
\end{table}
We evaluate the monomial selection methods on Rotated-MNIST, a dataset for hand-written digit recognition with randomly rotated inputs including 12k training and 50k testing grayscale-images \cite{RotMNIST}. Therefore, we train a SF-CNN with five convolutional and three fully connected layers where we insert II in between \cite{SFCNN,Rath}. For all layers, we use $n_\alpha=16$ rotations. Table \ref{tab:MonomialSelection} shows the performance of the different monomial selection algorithms.
We perform five runs for each dataset size using data augmentation with random rotations and report the mean test error and the standard deviation for the full dataset.

The results in Table \ref{tab:MonomialSelection} indicate that magnitude-based pruning with random pre-selection outperforms both the LSE baseline and the connectivity-pruning approach for monomial selection. Random initial selection outperforms the catalog-based approach. Furthermore, it is evident that the monomial selection algorithm plays a key part and allows a relative performance increase of up to $10.9 \%$ compared to a purely random monomial selection. Therefore, we use randomly initialized magnitude-based pruning with pre-training for all following monomial experiments.

\subsection{Evaluating Alternatives to Monomials on Digits}
 \begin{table*}
	\centering
	\caption{MTE on limited subsets of Rotated-MNIST using SF-CNN as baseline \cite{SFCNN}.}\label{tab:RotMNISTLimited}
	\small
	\begin{tabular}{ccc|cccccc}
		\hline
		& & & \multicolumn{6}{|c}{Number/Percentage of samples} \\
		G-Conv & II & $f$ & 500/4.2\% 
		& 1k/8.3\% & 2k/17\% & 4k/33\% & 6k/50\% & 12k/100\% \\
		\hline
		x & x & - & 8.635 & 
		7.205 & 5.586 & 4.684 & 4.324 & 3.664 $\pm$ 0.082 \\
		\hline 
		$\checkmark$ & x & Pooling & 3.543 & 
		2.529 & 1.660 & 1.337 & 1.126 & 0.714 $\pm$ 0.022 \\
		$\checkmark$ & $\checkmark$ & Monomials & 3.115 & 
		2.194 & 1.593 & 1.322 & 1.068 & 0.677 $\pm$ 0.031  \\
		$\checkmark$ & $\checkmark$ & Global-WS & 3.120 & 
		2.294 & 1.614 & 1.200 & 1.004 & 0.712 $\pm$ 0.027 \\
		$\checkmark$ & $\checkmark$ & Local-WS & 3.168 & 
		2.292 & 1.612 & 1.186 & 1.032 & 0.688 $\pm$ 0.032 \\
		$\checkmark$ & $\checkmark$ & MLP & 3.250 & 
		2.310 & 1.652 & 1.242 & 1.024 & 0.732 $\pm$ 0.023 \\
		$\checkmark$ & $\checkmark$ & MH-SA & 3.178 & 
		2.268 & 1.666 & 1.294 & 1.038 & 0.710 $\pm$ 0.022 \\
		$\checkmark$ & $\checkmark$ & SF-II & 3.352 & 
		2.542 & 1.836 & 1.346 & 1.128 & 0.782 $\pm$ 0.012 \\
		\hline
		\multicolumn{5}{l}{E(2)-CNN, Rotation} & & & & 0.705 $\pm$ 0.025 \\
		\multicolumn{5}{l}{E(2)-CNN, Rotation \& Flips} & & & & 0.682 $\pm$ 0.022 \\
	\end{tabular}
\end{table*}
\begin{table*}
	\small
	\centering
	\caption{MTE on limited subsets of SVHN using WRN16-4 \cite{WideResNet} as baseline.}\label{tab:SVHNLimited}
	\begin{tabular}{ccc|ccccc|c}
		\hline
		& & & \multicolumn{5}{|c|}{Number/Percentage of samples} & \\
		G-Conv & II & $f$ & 1k/1.3\% & 5k/6.9\% & 10k/14\% & 50k/69\% & 73k/100\% & \# Param. \\
		\hline
		x & x & - & 12.72 & 6.37 & 4.96 & 3.29 & 3.00 $\pm$ 0.01 & 2.75M \\
		\hline
		$\checkmark$ & x & Pooling & 11.15 & 5.52 & 4.46 & 3.25 & 2.89 $\pm$ 0.09 & 2.76M \\
		$\checkmark$ & $\checkmark$ & Monomials & \textbf{10.67} & 5.45 & 4.51 & 3.10 & 2.79 $\pm$ 0.03 & 2.78M\\
		$\checkmark$ & $\checkmark$ & Global-WS & 11.37 & 6.45 & 4.96 & 3.32 & 2.95 $\pm$ 0.07 & 2.83M \\
		$\checkmark$ & $\checkmark$ & Local-WS & \textbf{10.70} & \textbf{5.04} & \textbf{4.31} & \textbf{3.00} & \textbf{2.69} $\pm$ 0.01 & 2.77M \\
	\end{tabular}
\end{table*}
We further use Rotated-MNIST to evaluate the monomial replacement candidates using the training setup from above on full and limited datasets (Table \ref{tab:RotMNISTLimited}). We observe that all variants of II outperform the baseline SF-CNN utilizing pooling and a standard seven layer CNN trained with data augmentation (as used for comparison in \cite{GroupEquivariantCNNs}).
Especially in the limited-data domain, II-enhanced networks achieve a better performance despite adding more parameters. Consequently, II successfully reduces the data-complexity and thereby improves the generalization ability.
We conjecture that this is due to the II layer better preserving information that effectively contributes to successful classification compared to spatial pooling, i.e., II explicitly enforces invariance without afflicting other relevant information. 

For all practical purposes, monomial-based II performs on par with the alternative functions which enable a streamlined training procedure. Thus, it seems possible to replace the monomials with other functions in order to avoid the monomial selection step while maintaining the performance. This would further reduce the training time and at the same time provide a setup in which the II layer can be optimally tuned and adapted to the other layers in the network. All proposed functions are well-known in deep learning literature
which supports the practical deployment. In order to show that the benefits of II do not only stem from additional model capacity but from effectively leveraging prior knowledge, we add another steerable G-Conv and perform average pooling as a special case of II (SF-II) which performs clearly inferior. 

We outperform the E(2)-CNNs \cite{E2STCNNs} when they only incorporate invariance to rotations and achieve comparable results when they use a bigger invariance group including flips. The WS approach shows the most promising results among the different monomial replacement candidates. 

We also conduct experiments on SVHN in order to assess the performance of II on real-world datasets that do not involve artificially induced global invariances. It contains 73k training and 10k test samples of single digits from house numbers in its core dataset \cite{SVHN}. We use WRN16-4 as baseline \cite{WideResNet} and conduct experiments on the full dataset and limited subsets 
(Table \ref{tab:SVHNLimited}).
We compare the WRN to a SF-WRN and to II based on monomials, global- and local-WS with $k=3$.
For all following experiments, we use $n_\alpha=8$ angles for the steerable convolutions as well as II and perform three runs per network and dataset size. 

The II-based approach generally outperforms both the standard WRN16-4 as well as the equivariant baseline which achieves invariance using pooling. This proves that II is useful for real-world setups with non-transformed input data and can be applied to complex DNN architectures such as WRNs. The monomial and local-WS approach seem to perform best among all dataset sizes, with local-WS achieving slightly better results. We believe this is due to the fact that the architecture using this newly proposed function can be trained more efficiently. Additionally, training can be conducted in a single run without intermediate pruning steps since the monomial selection is avoided. 
Global-WS achieves worse results over all dataset sizes. Generally we assume that differences in performance among various methods over data size have to do with the trade-off between how good a specific architecture is able to leverage the prior knowledge on rotation invariance and how good it is able to learn and preserve other relevant invariance cues contained in real-world datasets such as color changes or illuminations.

\subsection{Object Classification on Real-World Natural Images}
\begin{table*}[t]
	\small
	\centering
	\caption{MTE on limited subsets of CIFAR-10 using WRN28-10 	\cite{WideResNet} as baseline.}\label{tab:CIFAR-Limited}
	\begin{tabular}{ccc|cccc|c}
		\hline
		& & & \multicolumn{4}{|c|}{Number/Percentage of samples} & \\
		G-Conv & II & $f$ & 100/0.2\% & 1k/2\% & 10k/20\% & 50k/100\% & \#Param. \\
		\hline
		x & x & - & 71.69 & 37.61 &\textbf{9.08} & \textbf{3.89} $\pm$ 0.02 & 36.5M \\
		\hline
		$\checkmark$ & x & Pooling & 76.54 & 37.29 & 12.68 & 4.71 $\pm$ 0.04 & 36.7M \\
		$\checkmark$ & $\checkmark$ & Monomials & \textbf{69.42} & \textbf{29.83} & 11.15 & 4.60 $\pm$ 0.12 & 36.8M \\
		$\checkmark$ & $\checkmark$ & Local-WS & 72.72 & 32.10 & 10.45 & 4.54 $\pm$ 0.15 & 36.9M \\
		\hline
		\multicolumn{3}{l}{E(2)-WRN28-10} & & & & \textbf{\textit{2.91}} & $\sim$37M \\
	\end{tabular}
\end{table*}
\begin{table}[t]
	\centering
	\small
	\caption{MTE on STL-10 using WRN16-8 \cite{WideResNet} as baseline.}\label{tab:STL-10}
	\begin{tabular}{ccccc}
		\hline
		G-Conv & II & $f$ & MTE[\%] & \# Param. \\
		\hline
		x & x & - & 12.74 $\pm$ 0.23 & 10.97M \\
		\hline
		$\checkmark$ & x & Pooling & 12.51 $\pm$ 0.33 & 10.83M \\
		$\checkmark$ & $\checkmark$ & Monomials & 10.84 $\pm$ 0.46 & 10.85M\\
		$\checkmark$ & $\checkmark$ & Local-WS & \textbf{10.09} $\pm$ 0.21 & 10.92M \\
		\hline
		\multicolumn{3}{l}{E(2)-WRN16-8} & 9.80 $\pm$ 0.40 & 12.0M \\
		\multicolumn{3}{l}{SES-WRN16-8} & \textbf{\textit{8.51}} & 11.0M \\ 
	\end{tabular}
\end{table}
To evaluate our approach on more complex classification settings including more variability, we use CIFAR-10 and STL-10. CIFAR-10 is an object classification dataset with 50k training and 10k test RGB-images \cite{CIFAR-10}. STL-10 is a subset of ImageNet containing 5,000 labeled training images from 10 classes \cite{STL10}. It is commonly used as a benchmark for semi-supervised learning and classification with limited training data. 

We use WRN28-10 and WRN16-8 as baseline architecture, respectively and test II with monomials and local-WS with $k=3$. For CIFAR-10, we train on full data as well as on limited subsets using standard data augmentation with random crops and flips (Table \ref{tab:CIFAR-Limited}). For STL-10 (Table \ref{tab:STL-10}), we use random crops, flips and cutout \cite{Cutout}.

On CIFAR-10, we notice two developments: While our networks outperform the WRN28-10 in the limited-data domain, indicating an improved sample complexity, they are unable to achieve better results in large-data regimes (Table \ref{tab:CIFAR-Limited}). Networks employing II achieve a better performance than the pooling counterpart among all dataset sizes indicating that II better preserves the information needed for a successful classification leading to a lower sample complexity. 

Local-WS performs on par or slightly worse than the monomials. We conjecture that on bigger dataset sizes, our approach with its rotation-invariant focus does not capture the complex local object-related invariant cues needed for successful classification as good as a standard WRN. We remark that for SVHN, relevant invariance cues besides rotation are rather global (e.g., color, illumination, noise), while for CIFAR these are also local and object-related (e.g., perspective changes, occlusions). Thus, our method handles global invariances well while needing additional steps to handle local invariances other than rotation. 

\cite{E2STCNNs} (E(2)-WRN) achieve better results than our networks in this setup. However, their approach differs from ours by loosening equivariance restrictions with depth and using a bigger invariance group including flips, thus addressing more local invariances. Nevertheless, this approach can be combined with ours in the future.

On STL-10, both II-enhanced networks outperform the equivariant baseline using pooling and the standard WRN. The local-WS approach outperforms the monomial counterpart. On this basis, we conclude that for all practical purposes, II based on local-WS delivers best results while being simpler to train than the monomial variant.
Again, other methods incorporating invariance to other groups such as the general E(2)-CNN \cite{E2STCNNs} or scales (SES-CNN, \cite{Sosnovik}) achieve better results than our purely rotation-invariant network. This is intuitive since samples from ImageNet involve variability from an even greater source of different transformations than CIFAR-10. Consequently, the invariance cues that need to be captured by a classifier are even more complex.

\section{CONCLUSION}\label{sec:Conclusion}
In this contribution, we focused on leveraging prior knowledge about invariance to transformations for classification problems. Therefore, we adapted the II framework by introducing a novel monomial selection algorithm and replacing the monomials with different functions such as a weighted sum, a MLP, and self-attention. Replacing the monomials enabled a streamlined training of DNNs using II by avoiding the pre-training and selection step. This allows to optimally tune and adapt all algorithmic components at once promoting the application of II to complex real-world datasets and architectures, e.g., WRNs. 

Our method explicitly enforces invariance which we see among the key factors to be taken into consideration by a feature-extraction engine for successful classification, especially for real-world applications, where data is often limited. Assuming that rotation invariance is required, we have shown how to design a DNN based on II to leverage this prior knowledge. In comparison to the standard approach, we replace spatial max-pooling by a dedicated layer which explicitly enforces invariance while increasing the network's expressibility. To enable the network to capture other invariance cues in particular of global nature we use a trainable weights as well.

We have demonstrated state-of-the-art sample complexity on datasets from various real-world setups. We achieve state-of-the-art results on all data regimes on image classification tasks when the targeted invariances (i.e., rotation) generate the most intraclass variance, as in the case of Rotated-MNIST and SVHN. On Rotated-MNIST, we even outperform the E(2)-CNN which also includes invariance to flips. 

On CIFAR-10 and STL-10, we show top performance in limited-data regimes for image classification tasks where various other transformations besides rotation are responsible for the intraclass variance. At the same time, the performance in the full-data regime is better than the equivariant baseline, which shows that we are able to effectively make use of prior knowledge and introduce rotation invariance without afflicting other learned invariances. Specifically, monomials and local-WS achieve the best and most stable performance and consistently outperform the baseline, which uses group and spatial pooling, as well as standard convolutional architectures. Local-WS performs similarly or better than monomials while being easier to apply and optimize due to avoiding the monomial selection step. It is different to simply adding an additional group-equivariant layer and performing average pooling among rotations and spatial locations because group pooling is performed before applying the II layer. Compared to TI-Pooling \cite{Laptev}, our method explicitly guarantees invariance within a single forward pass. In contrast, TI-Pooling approximates invariance by pooling among the responses of a non-equivariant network needing one forward pass per group element. 

Our current method is limited to problem settings where rotation invariance is desired. The expansion to other transformations is interesting future work. We also plan to investigate replacing all pooling operations with II.
\section*{\uppercase{Acknowledgements}}
The authors would like to thank their colleagues Lukas Enderich, Julia Lust and Paul Wimmer for their valuable contributions and fruitful discussions.
\bibliographystyle{apalike}
{\small
\bibliography{example}}

\begin{thebibliography}{}

\bibitem[Bekkers, 2020]{BekkersLie}
Bekkers, E.~J. (2020).
\newblock B-spline cnns on lie groups.
\newblock In {\em {ICLR} 2020, Addis Ababa, Ethiopia, April 26-30, 2020}.
  OpenReview.net.

\bibitem[Bekkers et~al., 2018]{Bekkers}
Bekkers, E.~J., Lafarge, M.~W., Veta, M., Eppenhof, K. A.~J., Pluim, J. P.~W.,
  and Duits, R. (2018).
\newblock Roto-translation covariant convolutional networks for medical image
  analysis.
\newblock In {\em {MICCAI} 2018, Granada, Spain, September 16-20, 2018,
  Proceedings, Part {I}}, pages 440--448.

\bibitem[Coates et~al., 2011]{STL10}
Coates, A., Ng, A.~Y., and Lee, H. (2011).
\newblock An analysis of single-layer networks in unsupervised feature
  learning.
\newblock In Gordon, G.~J., Dunson, D.~B., and Dud{\'{\i}}k, M., editors, {\em
  {AISTATS} 2011, Fort Lauderdale, USA, April 11-13, 2011}, volume~15 of {\em
  {JMLR} Proceedings}, pages 215--223. JMLR.org.

\bibitem[Cohen et~al., 2019a]{Gauge}
Cohen, T., Weiler, M., Kicanaoglu, B., and Welling, M. (2019a).
\newblock Gauge equivariant convolutional networks and the icosahedral {CNN}.
\newblock In {\em {ICML} 2019, 9-15 June 2019, Long Beach, CA, {USA}}, pages
  1321--1330.

\bibitem[Cohen and Welling, 2016]{GroupEquivariantCNNs}
Cohen, T. and Welling, M. (2016).
\newblock Group equivariant convolutional networks.
\newblock In {\em {ICML} 2016, New York City, NY, USA, June 19-24, 2016}, pages
  2990--2999.

\bibitem[Cohen et~al., 2019b]{GeneralEquivariantCNNs}
Cohen, T.~S., Geiger, M., and Weiler, M. (2019b).
\newblock A general theory of equivariant cnns on homogeneous spaces.
\newblock In Wallach, H.~M., Larochelle, H., Beygelzimer, A.,
  d'Alch{\'{e}}{-}Buc, F., Fox, E.~B., and Garnett, R., editors, {\em NeurIPS
  2019, 8-14 December 2019, Vancouver, BC, Canada}, pages 9142--9153.

\bibitem[Condurache and Mertins, 2012]{Condurache}
Condurache, A.~P. and Mertins, A. (2012).
\newblock Sparse representations and invariant sequence-feature extraction for
  event detection.
\newblock {\em VISAPP 2012}, 1.

\bibitem[Devries and Taylor, 2017]{Cutout}
Devries, T. and Taylor, G.~W. (2017).
\newblock Improved regularization of convolutional neural networks with cutout.
\newblock {\em CoRR}, abs/1708.04552.

\bibitem[Diaconu and Worrall, 2019a]{Diaconu2}
Diaconu, N. and Worrall, D.~E. (2019a).
\newblock Affine self convolution.
\newblock {\em CoRR}, abs/1911.07704.

\bibitem[Diaconu and Worrall, 2019b]{Diaconu1}
Diaconu, N. and Worrall, D.~E. (2019b).
\newblock Learning to convolve: {A} generalized weight-tying approach.
\newblock In {\em {ICML} 2019, 9-15 June 2019, Long Beach, California, {USA}},
  pages 1586--1595.

\bibitem[Esteves et~al., 2018a]{Esteves}
Esteves, C., Allen{-}Blanchette, C., Makadia, A., and Daniilidis, K. (2018a).
\newblock Learning {SO(3)} equivariant representations with spherical cnns.
\newblock In {\em {ECCV} 2018, Munich, Germany, September 8-14, 2018,
  Proceedings, Part {XIII}}, pages 54--70.

\bibitem[Esteves et~al., 2018b]{PolarTN}
Esteves, C., Allen-Blanchette, C., Zhou, X., and Daniilidis, K. (2018b).
\newblock Polar transformer networks.
\newblock In {\em ICLR 2018}.

\bibitem[Falkner et~al., 2018]{BOHB}
Falkner, S., Klein, A., and Hutter, F. (2018).
\newblock {BOHB}: Robust and efficient hyperparameter optimization at scale.
\newblock In {\em Proceedings of the 35th International Conference on Machine
  Learning}, pages 1436--1445.

\bibitem[Finzi et~al., 2021]{Finzi2021}
Finzi, M., Welling, M., and Wilson, A.~G. (2021).
\newblock A practical method for constructing equivariant multilayer
  perceptrons for arbitrary matrix groups.
\newblock {\em CoRR}, abs/2104.09459.

\bibitem[Freeman and Adelson, 1991]{Adelson}
Freeman, W.~T. and Adelson, E.~H. (1991).
\newblock The design and use of steerable filters.
\newblock {\em {IEEE} Trans. Pattern Anal. Mach. Intell.}, 13(9):891--906.

\bibitem[Fuchs et~al., 2020]{SE3Transformer}
Fuchs, F., Worrall, D.~E., Fischer, V., and Welling, M. (2020).
\newblock Se(3)-transformers: 3d roto-translation equivariant attention
  networks.
\newblock In Larochelle, H., Ranzato, M., Hadsell, R., Balcan, M., and Lin, H.,
  editors, {\em NeurIPS 2020, December 6-12, 2020, virtual}.

\bibitem[Fuchs et~al., 2021]{IterativeSE3Transformer}
Fuchs, F.~B., Wagstaff, E., Dauparas, J., and Posner, I. (2021).
\newblock Iterative se(3)-transformers.
\newblock {\em CoRR}, abs/2102.13419.

\bibitem[Ghosh and Gupta, 2019]{GhoshScale}
Ghosh, R. and Gupta, A.~K. (2019).
\newblock Scale steerable filters for locally scale-invariant convolutional
  neural networks.
\newblock {\em CoRR}, abs/1906.03861.

\bibitem[Han et~al., 2015]{MagnitudePruning}
Han, S., Pool, J., Tran, J., and Dally, W.~J. (2015).
\newblock Learning both weights and connections for efficient neural network.
\newblock In Cortes, C., Lawrence, N.~D., Lee, D.~D., Sugiyama, M., and
  Garnett, R., editors, {\em NeurIPS 2015, December 7-12, 2015, Montreal,
  Quebec, Canada}, pages 1135--1143.

\bibitem[Hutchinson et~al., 2020]{LieTransformer}
Hutchinson, M., Lan, C.~L., Zaidi, S., Dupont, E., Teh, Y.~W., and Kim, H.
  (2020).
\newblock Lietransformer: Equivariant self-attention for lie groups.
\newblock {\em CoRR}, abs/2012.10885.

\bibitem[Jaderberg et~al., 2015]{STN}
Jaderberg, M., Simonyan, K., Zisserman, A., and Kavukcuoglu, K. (2015).
\newblock Spatial transformer networks.
\newblock In {\em NeurIPS 2015}, pages 2017--2025. Curran Associates, Inc.

\bibitem[Kanazawa et~al., 2014]{LocalScaleInvariance}
Kanazawa, A., Sharma, A., and Jacobs, D.~W. (2014).
\newblock Locally scale-invariant convolutional neural networks.
\newblock {\em CoRR}, abs/1412.5104.

\bibitem[Kingma and Ba, 2015]{Adam}
Kingma, D.~P. and Ba, J. (2015).
\newblock Adam: {A} method for stochastic optimization.
\newblock In Bengio, Y. and LeCun, Y., editors, {\em {ICLR} 2015, San Diego,
  CA, USA, May 7-9, 2015, Conference Track Proceedings}.

\bibitem[Kondor et~al., 2018]{ClebschGordanNets}
Kondor, R., Lin, Z., and Trivedi, S. (2018).
\newblock Clebsch-gordan nets: a fully fourier space spherical convolutional
  neural network.
\newblock In Bengio, S., Wallach, H.~M., Larochelle, H., Grauman, K.,
  Cesa{-}Bianchi, N., and Garnett, R., editors, {\em NeurIPS 2018, 3-8 December
  2018, Montr{\'{e}}al, Canada}, pages 10138--10147.

\bibitem[Krizhevsky, 2009]{CIFAR-10}
Krizhevsky, A. (2009).
\newblock Learning multiple layers of features from tiny images,.
\newblock Technical report.

\bibitem[Laptev et~al., 2016]{Laptev}
Laptev, D., Savinov, N., Buhmann, J.~M., and Pollefeys, M. (2016).
\newblock {TI-POOLING:} transformation-invariant pooling for feature learning
  in convolutional neural networks.
\newblock In {\em {CVPR} 2016, Las Vegas, NV, USA, June 27-30, 2016}, pages
  289--297.

\bibitem[Larochelle et~al., 2007]{RotMNIST}
Larochelle, H., Erhan, D., Courville, A.~C., Bergstra, J., and Bengio, Y.
  (2007).
\newblock An empirical evaluation of deep architectures on problems with many
  factors of variation.
\newblock In {\em {ICML} 2007, Corvallis, Oregon, USA, June 20-24, 2007}, pages
  473--480.

\bibitem[LeCun et~al., 2015]{DeepLearning}
LeCun, Y., Bengio, Y., and Hinton, G.~E. (2015).
\newblock Deep learning.
\newblock {\em Nature}, 521(7553):436--444.

\bibitem[Lee et~al., 2019]{SNIP}
Lee, N., Ajanthan, T., and Torr, P. H.~S. (2019).
\newblock Snip: single-shot network pruning based on connection sensitivity.
\newblock In {\em {ICLR} 2019, New Orleans, LA, USA, May 6-9, 2019}.
  OpenReview.net.

\bibitem[Marcos et~al., 2018]{MarcosScale}
Marcos, D., Kellenberger, B., Lobry, S., and Tuia, D. (2018).
\newblock Scale equivariance in cnns with vector fields.
\newblock {\em CoRR}, abs/1807.11783.

\bibitem[Marcos et~al., 2017]{MarcosRot}
Marcos, D., Volpi, M., Komodakis, N., and Tuia, D. (2017).
\newblock Rotation equivariant vector field networks.
\newblock In {\em {ICCV} 2017, Venice, Italy, October 22-29, 2017}, pages
  5058--5067. {IEEE} Computer Society.

\bibitem[M{\"{u}}ller and Mertins, 2009]{Muller1}
M{\"{u}}ller, F. and Mertins, A. (2009).
\newblock Invariant-integration method for robust feature extraction in
  speaker-independent speech recognition.
\newblock In {\em {INTERSPEECH} 2009, Brighton, United Kingdom, September 6-10,
  2009}, pages 2975--2978.

\bibitem[M{\"{u}}ller and Mertins, 2010]{Muller2}
M{\"{u}}ller, F. and Mertins, A. (2010).
\newblock Invariant integration features combined with speaker-adaptation
  methods.
\newblock In {\em {INTERSPEECH} 2010, Makuhari, Chiba, Japan, September 26-30,
  2010}, pages 2622--2625.

\bibitem[M{\"{u}}ller and Mertins, 2011]{Muller3}
M{\"{u}}ller, F. and Mertins, A. (2011).
\newblock Contextual invariant-integration features for improved
  speaker-independent speech recognition.
\newblock {\em Speech Communication}, 53(6):830--841.

\bibitem[Netzer et~al., 2011]{SVHN}
Netzer, Y., Wang, T., Coates, A., Bissacco, A., Wu, B., and Ng, A.~Y. (2011).
\newblock Reading digits in natural images with unsupervised feature learning.
\newblock {\em NIPS Workshop on Deep Learning and Unsupervised Feature
  Learning}.

\bibitem[Noether, 1916]{Noether1916}
Noether, E. (1916).
\newblock Der endlichkeitssatz der invarianten endlicher gruppen.
\newblock {\em Mathematische Annalen}, 77:89--92.

\bibitem[Rath and Condurache, 2020]{Rath}
Rath, M. and Condurache, A.~P. (2020).
\newblock Invariant integration in deep convolutional feature space.
\newblock In {\em {ESANN} 2020, Bruges, Belgium, October 2-4, 2020}, pages
  103--108.

\bibitem[Reisert and Burkhardt, 2006]{II3d}
Reisert, M. and Burkhardt, H. (2006).
\newblock Invariant features for 3d-data based on group integration using
  directional information and spherical harmonic expansion.
\newblock In {\em {ICPR} 2006, 20-24 August 2006, Hong Kong, China}, pages
  206--209. {IEEE} Computer Society.

\bibitem[Romero et~al., 2020]{AttentiveGCNN}
Romero, D.~W., Bekkers, E.~J., Tomczak, J.~M., and Hoogendoorn, M. (2020).
\newblock Attentive group equivariant convolutional networks.
\newblock In {\em {ICML} 2020, 13-18 July 2020, Virtual Event}, pages
  8188--8199. {PMLR}.

\bibitem[Romero and Cordonnier, 2020]{G-SelfAttention}
Romero, D.~W. and Cordonnier, J. (2020).
\newblock Group equivariant stand-alone self-attention for vision.
\newblock {\em CoRR}, abs/2010.00977.

\bibitem[Romero and Hoogendoorn, 2020]{RomeroCoAttentive}
Romero, D.~W. and Hoogendoorn, M. (2020).
\newblock Co-attentive equivariant neural networks: Focusing equivariance on
  transformations co-occurring in data.
\newblock In {\em {ICLR} 2020, Addis Ababa, Ethiopia, April 26-30, 2020}.
  OpenReview.net.

\bibitem[Schulz-Mirbach, 1992]{SM_Existence}
Schulz-Mirbach, H. (1992).
\newblock On the existence of complete invariant feature spaces in pattern
  recognition.
\newblock In {\em Pattern Recognition: Eleventh International Conference 1992},
  pages 178 -- 182.

\bibitem[Schulz-Mirbach, 1994]{SM_Algos}
Schulz-Mirbach, H. (1994).
\newblock Algorithms for the construction of invariant features.
\newblock In {\em Tagungsband Mustererkennung 1994 (16. DAGM Symposium), Reihe
  Informatik Xpress, Nr.5}, pages 324--332.

\bibitem[Schulz{-}Mirbach, 1995]{SM_Gray}
Schulz{-}Mirbach, H. (1995).
\newblock Invariant features for gray scale images.
\newblock In {\em Mustererkennung 1995, 17. DAGM-Symposium, Bielefeld, 13.-15.
  September 1995, Proceedings}, pages 1--14.

\bibitem[Shaw et~al., 2018]{PeterShaw}
Shaw, P., Uszkoreit, J., and Vaswani, A. (2018).
\newblock Self-attention with relative position representations.
\newblock In Walker, M.~A., Ji, H., and Stent, A., editors, {\em NAACL-HLT
  2018, New Orleans, Louisiana, USA, June 1-6, 2018, Volume 2 (Short Papers)},
  pages 464--468. Association for Computational Linguistics.

\bibitem[Sosnovik et~al., 2020]{Sosnovik}
Sosnovik, I., Szmaja, M., and Smeulders, A. W.~M. (2020).
\newblock Scale-equivariant steerable networks.
\newblock In {\em {ICLR} 2020, Addis Ababa, Ethiopia, April 26-30, 2020}.

\bibitem[Tai et~al., 2019]{ETN}
Tai, K.~S., Bailis, P., and Valiant, G. (2019).
\newblock Equivariant transformer networks.
\newblock In {\em {ICML} 2019, 9-15 June 2019, Long Beach, California, {USA}},
  pages 6086--6095.

\bibitem[Veeling et~al., 2018]{PCAM}
Veeling, B.~S., Linmans, J., Winkens, J., Cohen, T., and Welling, M. (2018).
\newblock Rotation equivariant {CNNs} for digital pathology.
\newblock {\em CoRR}, abs/1806.03962.

\bibitem[Walters et~al., 2020]{ECCO}
Walters, R., Li, J., and Yu, R. (2020).
\newblock Trajectory prediction using equivariant continuous convolution.
\newblock {\em CoRR}, abs/2010.11344.

\bibitem[Weiler and Cesa, 2019]{E2STCNNs}
Weiler, M. and Cesa, G. (2019).
\newblock General e(2)-equivariant steerable cnns.
\newblock In {\em NeurIPS 2019, 8-14 December 2019, Vancouver, BC, Canada},
  pages 14334--14345.

\bibitem[Weiler et~al., 2018a]{3DSTCNNs}
Weiler, M., Geiger, M., Welling, M., Boomsma, W., and Cohen, T. (2018a).
\newblock 3d steerable cnns: Learning rotationally equivariant features in
  volumetric data.
\newblock In Bengio, S., Wallach, H.~M., Larochelle, H., Grauman, K.,
  Cesa{-}Bianchi, N., and Garnett, R., editors, {\em NeurIPS 2018, 3-8 December
  2018, Montr{\'{e}}al, Canada}, pages 10402--10413.

\bibitem[Weiler et~al., 2018b]{SFCNN}
Weiler, M., Hamprecht, F.~A., and Storath, M. (2018b).
\newblock Learning steerable filters for rotation equivariant cnns.
\newblock In {\em {CVPR} 2018, Salt Lake City, UT, USA, June 18-22, 2018},
  pages 849--858.

\bibitem[Winkels and Cohen, 2019]{Winkels}
Winkels, M. and Cohen, T.~S. (2019).
\newblock Pulmonary nodule detection in {CT} scans with equivariant cnns.
\newblock {\em Medical Image Anal.}, 55:15--26.

\bibitem[Worrall and Brostow, 2018]{CubeNet}
Worrall, D.~E. and Brostow, G.~J. (2018).
\newblock Cubenet: Equivariance to 3d rotation and translation.
\newblock In {\em {ECCV} 2018, Munich, Germany, September 8-14, 2018,
  Proceedings, Part {V}}, pages 585--602.

\bibitem[Worrall et~al., 2017]{HarmonicNetworks}
Worrall, D.~E., Garbin, S.~J., Turmukhambetov, D., and Brostow, G.~J. (2017).
\newblock Harmonic networks: Deep translation and rotation equivariance.
\newblock In {\em 2017 {IEEE} Conference on Computer Vision and Pattern
  Recognition, {CVPR} 2017, Honolulu, HI, USA, July 21-26, 2017}, pages
  7168--7177.

\bibitem[Worrall and Welling, 2019]{Worrall19}
Worrall, D.~E. and Welling, M. (2019).
\newblock Deep scale-spaces: Equivariance over scale.
\newblock In {\em NeurIPS 2019, 8-14 December 2019, Vancouver, BC, Canada},
  pages 7364--7376.

\bibitem[Xu et~al., 2014]{XuScale}
Xu, Y., Xiao, T., Zhang, J., Yang, K., and Zhang, Z. (2014).
\newblock Scale-invariant convolutional neural networks.
\newblock {\em CoRR}, abs/1411.6369.

\bibitem[Yang et~al., 2019]{RegularizedInvariance}
Yang, F., Wang, Z., and Heinze{-}Deml, C. (2019).
\newblock Invariance-inducing regularization using worst-case transformations
  suffices to boost accuracy and spatial robustness.
\newblock In {\em NeurIPS 2019, 8-14 December 2019, Vancouver, BC, Canada},
  pages 14757--14768.

\bibitem[Zagoruyko and Komodakis, 2016]{WideResNet}
Zagoruyko, S. and Komodakis, N. (2016).
\newblock Wide residual networks.
\newblock In Wilson, R.~C., Hancock, E.~R., and Smith, W. A.~P., editors, {\em
  {BMVC} 2016, York, UK, September 19-22, 2016}. {BMVA} Press.

\bibitem[Zhu et~al., 2019]{ZhuScale}
Zhu, W., Qiu, Q., Calderbank, A.~R., Sapiro, G., and Cheng, X. (2019).
\newblock Scale-equivariant neural networks with decomposed convolutional
  filters.
\newblock {\em CoRR}, abs/1909.11193.

\end{thebibliography}

\section*{\uppercase{Appendix}}

\begin{table*}
	\caption{II-SF-CNN hyper-parameters on Rotated-MNIST.}\label{tab:HPRotMNIST}
	\small
	\centering
	\begin{tabular}{cccccccc}
		Hyper-parameter & MH-SA & Global-WS & Local-WS & MLP & SF-Conv & Monomials & SF-CNN \\ 
		\hline 		
		Optimizer & Adam & Adam & Adam & Adam & Adam & Adam & Adam \\
		Batch Size & 32 & 32 & 32 & 32 & 32 & 32 & 64 \\
		Epochs & 100 & 100 & 100 & 100 & 100 & 100 & 100 \\
		$n_{FC}$ & 95 & 85 & 30 & 85 & 85 & 90 & 96\\
		Learning Rate &  5e-3 & 1e-4 & 1e-3 & 1e-4 & 5e-4 & 1e-4 & 1e-3 \\  
		LR Decay &  0.5 & 0.1 & 0.5 & 0.1  & 0.2 & 0.75 & 0.9 \\  
		LR Decay Epoch &  20 & 40 & 25 & 30 & 25 & 15 & 20 \\  
		Reg. Constant &  1e-3 & 0.1 & 1e-3 & 1e-3 & 0.01 & 0.15 & 1. \\ 
		Dropout Rate & 0.05 & 0.45 & 0.4 & 0.5 & 0.4 & 0.45 & 0.7\\ 
		Attention Heads & 1 & - & - & - & - & - & -\\
		Attention Dropout & 0. & - & - & - & - & - & - \\
	\end{tabular} 
\end{table*}

\begin{table*}
	\centering
	\caption{Hyper-parameters on SVHN.}\label{tab:HPSVHN}
	\small
	\begin{tabular}{cccccc}
		Hyper-parameter & SF-CNN & Global-WS & Local-WS & Monomials \\ 
		\hline
		Optimizer & Adam & Adam & Adam & Adam \\
		Batch Size & 128 & 128 & 128 & 64 \\
		Epochs & 100 & 100 & 100 & 100 \\
		$n_{FC}$ & 32 & 64 & 36 & 85 \\
		Learning Rate  &  1e-3 & 5e-4 & 5e-4 & 5e-4 \\
		LR Decay  &  0.4 &  0.1 & 0.25 & 0.25 \\  
		LR Decay Epoch  & 20  & 30 & 25 & 20\\  
		Reg. Constant  & 2e-3  & 0.25 & 0.2 & 0.05 \\ 
		Dropout Rate  & 0.55 & 0.7 & 0.5 & 0.7\\
	\end{tabular} 
\end{table*}

\begin{table}
	\caption{Hyper-parameters on CIFAR-10.}\label{tab:HPCIFAR}
	\centering
	\small
	\begin{tabular}{cccc}
		Hyper-parameter & SF-CNN & Local-WS & Monomials \\ 
		\hline 
		Optimizer & Adam & Adam & Adam \\
		Batch Size & 64 & 64 & 64 \\
		Epochs & 100 & 100 & 100 \\
		$n_{FC}$ & - & 90 & 30 \\
		Learning Rate& 1e-3 & 5e-4 & 5e-4 \\
		LR Decay &  0.5 & 0.2 &  0.025\\  
		LR Decay Epoch & 50 & 50 & 50 \\  
		Reg. Constant &  0.1 & 5e-6 & 0.008 \\ 
		Dropout Rate & 0.3 & 0.1 & 0.4 \\
	\end{tabular} 
\end{table}

\begin{table}
	\caption{Hyper-parameters on STL-10.}\label{tab:HPSTL}
	\centering
	\small
	\begin{tabular}{cccc}
		Hyper-parameter & SF-CNN & Local-WS & Monomials \\ 
		\hline 
		Optimizer & Adam & Adam & Adam \\
		Batch Size & 96 & 64 & 32 \\
		Epochs & 1000 & 1000 & 1000 \\
		$n_{FC}$ & - & 16 & 10 \\
		Learning Rate  & 5e-4 & 0.01 & 5e-4 \\
		LR Decay  &  0.1 & 0.3 &  0.1\\  
		LR Decay Epoch & 300 & 300 & 300 \\  
		Reg. Constant  &  1e-8 & 1e-9 & 5e-9 \\ 
		Dropout Rate  & 0.1 & 0.15 & 0.05 \\
	\end{tabular} 
\end{table}

\paragraph{Implementation Details}\label{sec:AppendixHPs}
To increase the reproducibility, we provide our exact hyper-parameter settings.
We optimized the standard Wide-ResNets using stochastic gradient descent and the hyper-parameters of the corresponding paper \cite{WideResNet}. All steerable networks were optimized using Adam optimization \cite{Adam}.  We used exponential learning rate decay for Rotated-MNIST and SVHN, while we employed step-wise decay on CIFAR-10 and STL-10. All steerable filter weights were regularized using elastic net regularization with factor $10^{-7}$ (cf. \cite{SFCNN}). For all WRNs, we additionally use $\ell_2$-regularization for the learnable BatchNorm coefficients with factor $0.1$. All regularization losses were then multiplied by the regularization constant. 

The hyper-parameters were optimized using Bayesian Optimization with Hyperband (BOHB, \cite{BOHB}) on 80/20 validation splits, if it was not already predetermined by the dataset. They are shown in Tables \ref{tab:HPRotMNIST}-\ref{tab:HPSTL}. On Rotated-MNIST we used data augmentation with random rotations following \cite{SFCNN}. On CIFAR-10 and SVHN, we followed \cite{WideResNet} and used random crops and flips for CIFAR-10 and no data augmentation for SVHN. On STL-10, we use random crops, flips and cutout \cite{Cutout} .

For the monomial architectures, we applied II per channel, and pruned $M=50$ initial monomials to $n_m=5$ for Rot-MNIST and $n_m=10$ monomials for SVHN, CIFAR-10 and STL-10. We used one intermediate pruning step after 10 epochs with $n_m = 25$ and train additional 5 epochs before the final pruning step. All other invariant integration layers were implemented with constant number of channels. 

\paragraph{Number of Parameters}\label{sec:AppendixNumber}
For our invariant architectures, we keep the number of parameters constant by reducing the number of channels accordingly.
A standard convolutional filter with kernel size $k$, $c_i$ input channels and $c_o$ output channels has $k^2c_ic_o$ parameters.  A rotation-steerable filter has $2n_Fn_\alpha c_ic_o$ parameters with $n_\alpha$ rotations and $n_F$ basis filters. In order to keep the number of parameters constant, we equate
\begin{equation}
\small
k^2c_ic_o = 2n_Fn_\alpha \tilde{c}_i\tilde{c}_o \quad \Leftrightarrow \quad \frac{\tilde{c}_i\tilde{c}_o}{c_ic_o} = \frac{2n_Fn_\alpha}{k^2}
\end{equation}
We use $k=3$, $n_\alpha=8$, $n_F=16$ and obtain a final ratio of $\frac{256}{9}$ by which we need to reduce $c_ic_o$. Hence, we reduce the number of channels by $\sqrt{\frac{256}{9}} = \frac{16}{3}$. For the lifting convolution, the filter is not used among all rotations, so we only need to reduce the ratio by $\sqrt{\frac{32}{9}}$.
\end{document}